\documentclass{article}
\usepackage[utf8]{inputenc}
\usepackage{graphicx}
\usepackage[square,numbers]{natbib}
\usepackage{xcolor}
\usepackage{booktabs}
\usepackage{tabulary}
\usepackage[margin=1.25in]{geometry}
\usepackage{pdflscape}
\usepackage{authblk}
\usepackage[pdfborder={0 0 0}]{hyperref}
\usepackage{wrapfig}
\usepackage{comment}
\usepackage{soul}
\usepackage{tikz}

\makeatletter
\newcommand{\printfnsymbol}[1]{%
  \textsuperscript{\@fnsymbol{#1}}%
}

\newcommand\copyrighttext{%
  \footnotesize \textcopyright 2021 IEEE. Personal use of this material is permitted.
  Permission from IEEE must be obtained for all other uses, in any current or future
  media, including reprinting/republishing this material for advertising or promotional
  purposes, creating new collective works, for resale or redistribution to servers or
  lists, or reuse of any copyrighted component of this work in other works.
  DOI: \href{https://ieeexplore.ieee.org/document/9681717}{10.1109/MRA.2021.3138383}}
\newcommand\copyrightnotice{%
\begin{tikzpicture}[remember picture,overlay]
\node[anchor=south,yshift=10pt] at (current page.south) {\fbox{\parbox{\dimexpr\textwidth-\fboxsep-\fboxrule\relax}{\copyrighttext}}};
\end{tikzpicture}%
}

\makeatother

\title{A Tool for Organizing Key Characteristics of Virtual, Augmented, and Mixed Reality for Human-Robot Interaction Systems: Synthesizing VAM-HRI Trends and Takeaways}
\author{Thomas R. Groechel \thanks{equal contribution}}
\affil{U. of Southern California \\
\url{groechel@usc.edu}}

\author{Michael E. Walker \printfnsymbol{1}}
\author{Christine T. Chang}
\affil{U. of Colorado Boulder
\url{{michael.walker-1, christine.chang}@colorado.edu}}

\author{Eric Rosen}
\author{Jessica Zosa Forde}
\affil{Brown University 
\url{{eric_rosen, jessica_forde}@brown.edu}}

\date{\vspace{-8ex}}

\begin{document}


\maketitle

\copyrightnotice 

\begin{abstract}


Frameworks have begun to emerge to categorize Virtual, Augmented, and Mixed Reality (VAM) technologies that provide immersive, intuitive interfaces to facilitate Human-Robot Interaction. These frameworks, however,  fail to capture key characteristics of the growing subfield of VAM-HRI and can be difficult to consistently apply due to continuous scales. This work builds upon these prior frameworks through the creation of a Tool for Organizing Key Characteristics of VAM-HRI Systems (TOKCS). TOKCS discretizes the continuous scales used within prior works for more consistent classification and adds additional characteristics related to a robot's internal model, anchor locations, manipulability, and the system's software and hardware. To showcase the tool's capability, TOKCS is applied to \textcolor{black}{the ten papers from the fourth VAM-HRI workshop and examined for key} trends and takeaways. These trends highlight the expressive capability of TOKCS while also helping frame newer trends and future work recommendations for VAM-HRI research.

\end{abstract}

\section{Introduction}



The need to help identify growing trends within Virtual, Augmented, and Mixed Reality for Human Robot Interaction (VAM-HRI) is evidenced by four consecutive years of a VAM-HRI workshop consistently spanning 60-100+ attendees. This nascent sub-field of HRI addresses challenges in mixed reality interactions between humans and robots, involving applications such as remote teleoperation, mental model alignment for effective partnering, facilitating robot learning, and comparing the capabilities and perceptions of robots and virtual agents. VAM-HRI research is becoming even more accessible to the robotics community due in part to the wide-spread availability of commercial virtual reality (VR), augmented reality (AR), and mixed reality (MR) platforms and the rise of readily-accessible 3D game engines for supporting virtual environment interactions.

To understand what challenges and solutions have been focused on by this new community, \citet{williams2019reality} proposed the Reality-Virtuality Interaction cube as a tool for clustering VAM-HRI research. The Interaction Cube is a three-dimensional conceptual framework that captures characteristics about the design elements involved (expressivity of the view and flexibility of control) as well as the virtuality they implement (from real to fully virtual). While the Interaction Cube provides a useful lens for roughly characterizing research involving interactive technologies within VAM-HRI, the continuous nature of the cube makes it challenging to exactly position where design elements and environments are within the cube. Furthermore, the Interaction cube does not address other characteristics of VAM-HRI research that have recently gained attention, such as robot internal models, software, hardware, and experimental evaluation methods.


To help advance the understanding of different VAM-HRI systems, we introduce a \textbf{Tool for Organizing Key Characteristics of VAM-HRI Systems (TOKCS)}. TOKCS builds off work from the Interaction Cube, discretizing its continuous scales and adding new key characteristics for classification. The tool is applied to the 10 workshop papers from the $4^{th}$ International Workshop on VAM-HRI to validate its usefulness within the growing subfield. These classifications help inform current and future trends found within the workshop \textcolor{black}{and VAM-HRI as a whole}.

\section{Interaction Cube Framework}

The Interaction Cube \cite{williams2019reality} uses three dimensions to characterize VAM-HRI work: the 2D Plane of Interaction to represent interactive design elements and the 1D Reality-Virtuality Continuum from Milgram \cite{milgram1995augmented} to characterize the environment.
\begin{wrapfigure}{C}{0.6\textwidth}
\centering
\includegraphics[width=90mm]{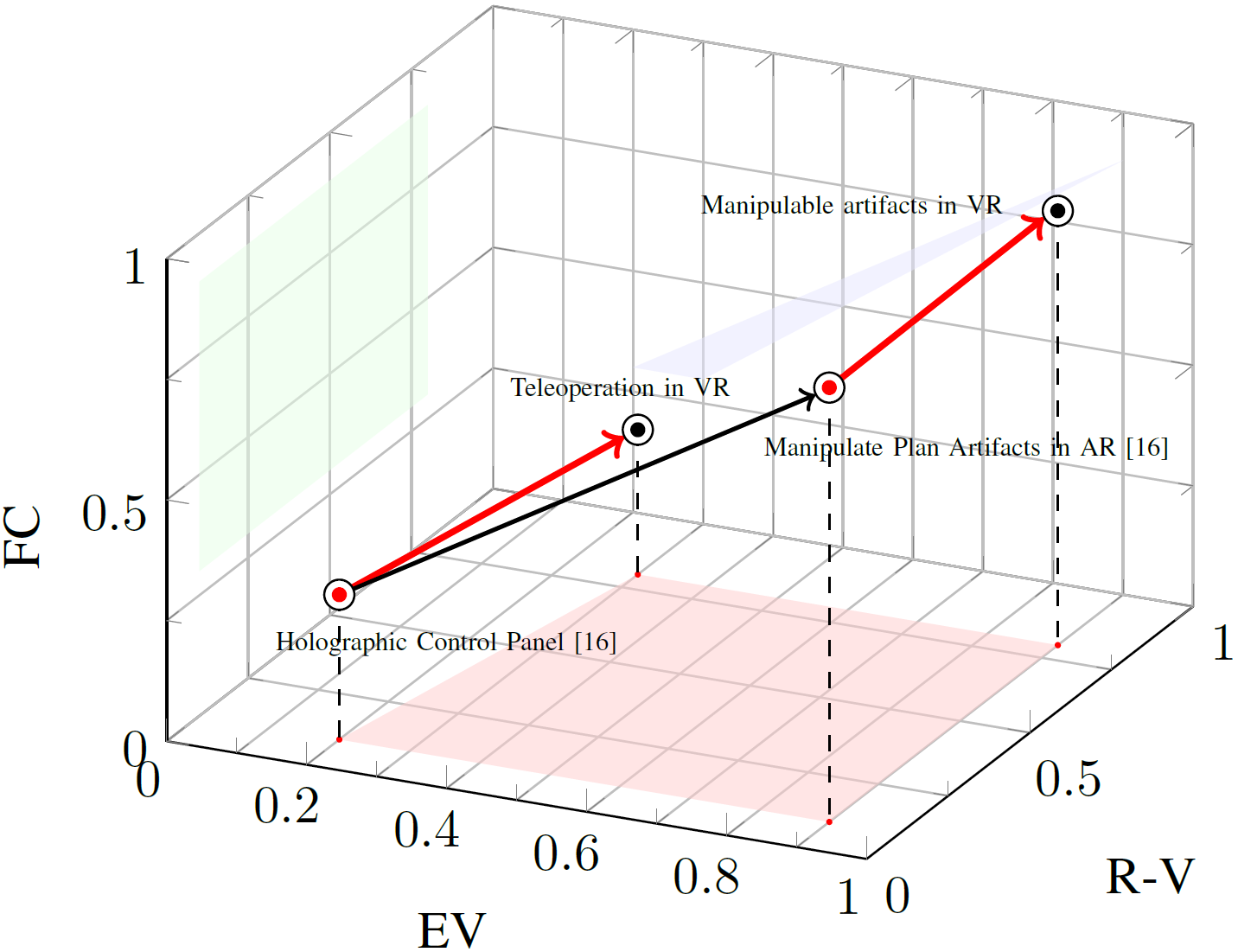}
\caption{The Reality-Virtuality Interaction Cube used to visually categorize MRIDEs according to their Flexibility of Control (FC), Expressivity of View (EV), and where they lie upon the Reality-Virtuality Continuum (RV). \textcolor{black}{Reality is indicated as 0 and Virtuality as 1.}}
\label{fig:Cube}
\end{wrapfigure}

\subsection{\textcolor{black}{Interaction Design Elements: Enhancing View and Control}}
\label{sec:interactiondesignel}
The first two dimensions of the Interaction Cube \textcolor{black}{(Fig. \ref{fig:Cube})} are defined by the \textit{Plane of Interaction}, which captures both (1) the opportunities to view into the robot's internal model, and (2) the degree of control the human has over the internal model. These two levels of interactivity (termed the \textbf{expressivity of view (EV)} and \textbf{flexibility of controller (FC)} respectively) are the conceptual pillars for characterizing interactivity within the Interaction Cube, and any components that contribute or impact either EV or FC are called \textit{interaction design elements}. This is similar to the Model-View-Controller design pattern. However, in this case the 2D placement on the Interaction Plane depends on a vector whose direction results from the impact a design element has on EV and the impact a design element has on FC. The magnitude of the vector is scaled by the complexity of the robot's internal model. According to \citet{williams2019reality}, ``while it is likely infeasible to explicitly determine the position of a technology on this plane, it is nevertheless instructive to consider the formal relationship between interaction design elements and the position of a technology on this plane.''

\subsection{\textcolor{black}{Mixed-Reality Interaction Design Elements: Anchoring and Artifacts}}
\label{sec:mrides}
 The Interaction Cube categorizes the study of VAM virtual objects as MRIDEs (mixed-reality interaction design elements), which can fall into one of three categories:\\
\begin{itemize}
  \item \textit{User-Anchored Interface Elements}: Objects attached to user view. This is similar to traditional GUI elements that are anchored to the user's \textcolor{black}{camera} coordinate frame and do not change along with the user's field of view. \textcolor{black}{These elements may also be referred to as part of a user's heads up display as popularized by video games and movies.}
  \item \textit{Environment-Anchored Interface Elements}: Objects anchored to the environment or robot. \textcolor{black}{For example, virtual arms that can be  anchored to a robot \cite{groechel2019using} or virtual objects that can be anchored to the physical environment.}
  \item \textit{Virtual Artifacts}: Objects that can be manipulated by humans or robots or may move ``under their own ostensible volition'' \cite{williams2019reality}. \textcolor{black}{For example, virtual indicators of robot position, such as arrows, can move on their own within the environment.}
\end{itemize}


\subsection{The Reality-Virtuality Continuum \& VAM-HRI} \label{sec:milgram}
The third axis of the Reality-Virtuality Interaction Cube illustrates where an MRIDE falls on the Reality-Virtuality Continuum \cite{milgram1995augmented}. This continuum classifies environments and interfaces with respect to how much virtual and/or real content they contain. On one end of the spectrum lies reality, which is any interface that does not use any virtual content and makes use of only real objects and imagery. The opposite end of the spectrum is virtual reality, which would be an interface that consists of pure virtual content without any integration of the real world (for example, a simulated world presented in VR). Between these two extremes is mixed reality, which captures all interfaces that incorporate a portion of both reality and virtuality in their design. There are two sub-classes of mixed reality: (1) augmented reality where virtual objects are integrated into the real world; and (2) augmented virutality where real objects are inserted within virtual environments.

\textbf{Augmented reality} interfaces in VAM-HRI often communicate the state and/or intentions of a real robot. For example, the battery levels of a robot can be displayed with a virtual object that hovers over a real robot, or a robot's planned trajectory can be drawn on the floor with a virtual line to indicate the robot's future movement intentions.

\textbf{Virtual reality} interfaces are often used to provide simulated environments where human users can interact with virtual robots. In these virtual settings user interactions with robots can be monitored and evaluated without risk of physical harm for either robot or human. Additionally, the virtual robot models can be easily and quickly altered to allow for rapid prototyping of both robot and interface design. Without the need for physical hardware, robots can be added to any virtual scene without the typical costs associated with real robots.

Virtual environments can also be used to teleoperate and/or supervise real robots in the physical world. In cases like these, 3D data collected by the real robot about its surrounding environment is integrated within virtual settings to create \textbf{augmented virtuality} interfaces. Cyber-physical interfaces and virtual control rooms are two common VAM-HRI augmented virtuality methods of enhancing remote robot operators ability by increasing situational awareness of their robot's state and location while mitigating the limitations of virtual interfaces such as cyber sickness \cite{lipton2017baxter}. 

\section{\textcolor{black}{The TOKCS Classification Framework}}
The key insight of this work is the addition of key characteristics of VAM-HRI not covered by the Interaction Cube to create TOKCS. These include VAM-HRI system hardware, research that seeks to increase the robot's model of the world around it, and additional granularity to \emph{mixed-reality} interaction design elements (MRIDEs). The characteristics are part of TOKCS which is then applied to the $4^{th}$ VAM-HRI workshop's papers in Sec. \ref{sec::paperclassifications}. The application informs the insights and future work recommendations outlined in Sec. \ref{sec::currenttrendsfuture}.

\subsection{Hardware} 
\label{sec:hardware}
While hardware used for virtual, augmented, and mixed reality can vary widely, there are certain types of hardware that are commonly used in VAM-HRI. Here we outline the most common, which enable experiences along the Reality-Virtuality Continuum: head-mounted displays (HMDs), projectors, displays, and peripherals. Because hardware technology is making significant advances every year, labeling the specific technology (e.g., HoloLens 2) is important when classifying hardware within TOKCS. These hardware technologies then fall under these categories.

\textbf{HMDs.} Virtual, mixed, and augmented reality all commonly use head-mounted displays. The Oculus Quest and HTC Vive both allow for a full virtual reality experience, visually immersing the user in a completely virtual environment. The HTC Vive also allows for augmented virtuality, such as in \citet{wadgaonkar2021exploring}, where the user is in a virtual setting but the virtual robot being manipulated is also moving in the real world. The Microsoft HoloLens and the Magic Leap are strictly augmented reality headsets, where virtual images are rendered on top of the real world view of the user.

\textbf{Projectors.} Onboard projectors can provide a way for the robot itself to display virtual objects or information. Alternately, static projectors allow an area to contain augmented reality elements. Images might be projected onto an object, on the floor, or onto a robot. 

\textbf{Displays.} This category of hardware ranges from handheld smartphones or tablets to room-size displays. Two-dimensional and three-dimensional monitors fall somewhere in between this range. Some of these exist in a single location, while mobile displays can be carried by a person or moved by a robot. A cave automated virtual environment (or CAVE) immerses the user in virtual reality using 3 to 6 walls to partially or fully enclose the space. An augmented reality display might include a realtime camera with overlaid virtual graphics, while a virtual reality display contains completely virtual graphics. Displays can be an especially effective way to conduct user studies without investing in expensive hardware, for example by showing recorded videos to participants on Amazon Mechanical Turk \cite{mott2021you}.

\textbf{Peripherals.} Peripheral devices allow for a richer interaction within virtual, augmented, or mixed reality. Leap Motion hand tracking can be combined with a headset such as the HTC Vive (as in \cite{mara2021cobot}) to provide recording and playback of motions and commands. Oculus Quest controllers are handheld and can be used individually or in tandem, giving the user a modality for both gesturing and selecting with the use of buttons on the device. Peripherals might frequently be used to enhance the Flexibility of Control (FC) of a MRIDE.


\textcolor{black}{\subsection{Software}
\label{sec:software}
There are a variety of software applications for facilitating 3D environments for VAM-HRI research. The most popular platforms like Unity3D support a wide variety of VR and MR hardware like those outlined in Section \ref{sec:hardware}, and offer packages for networking with robot networks like ROS servers and rendering robot sensor data. ROS also offers a robot simulator, Gazebo, that directly interfaces with ROS applications and which has been used for VAM-HRI research. Other additional software generally relevant to HRI research is also included here, such as tracking AR tags to detect object poses using TagUp \cite{Barentine2021-lu}. Software is not a direct part of the interaction as hardware, but we report relevant software for a holistic understanding of what resources the VAM-HRI community uses to develop their applications. }

\subsection{Robot Internal Complexity of Model}
\label{sec::modelcomplexity}
The Interaction Cube emphasizes the increased expressivitiy of view and flexibility of controller aspects of projected visual objects having on the robot's underlying model. This fails to explore, however, the sensing capabilities and data afforded by VAM technologies (e.g., ARHMD). The framework can be expanded by including the technologies' ability to aid the robot's internal model of the world - namely increasing the robot's internal \textbf{complexity of model (CM)}. The robot's internal CM benefits from data typically difficult to gather (e.g., eye-gaze) as well as the technology affording data assumptions (e.g., a headset with various sensors being anchored to the user's head). These data manifest in aiding a robot's model of the \textit{environment} and/or model of the \textit{user}. 

\begin{figure}[b!]
\centering
\includegraphics[width=0.69\textwidth]{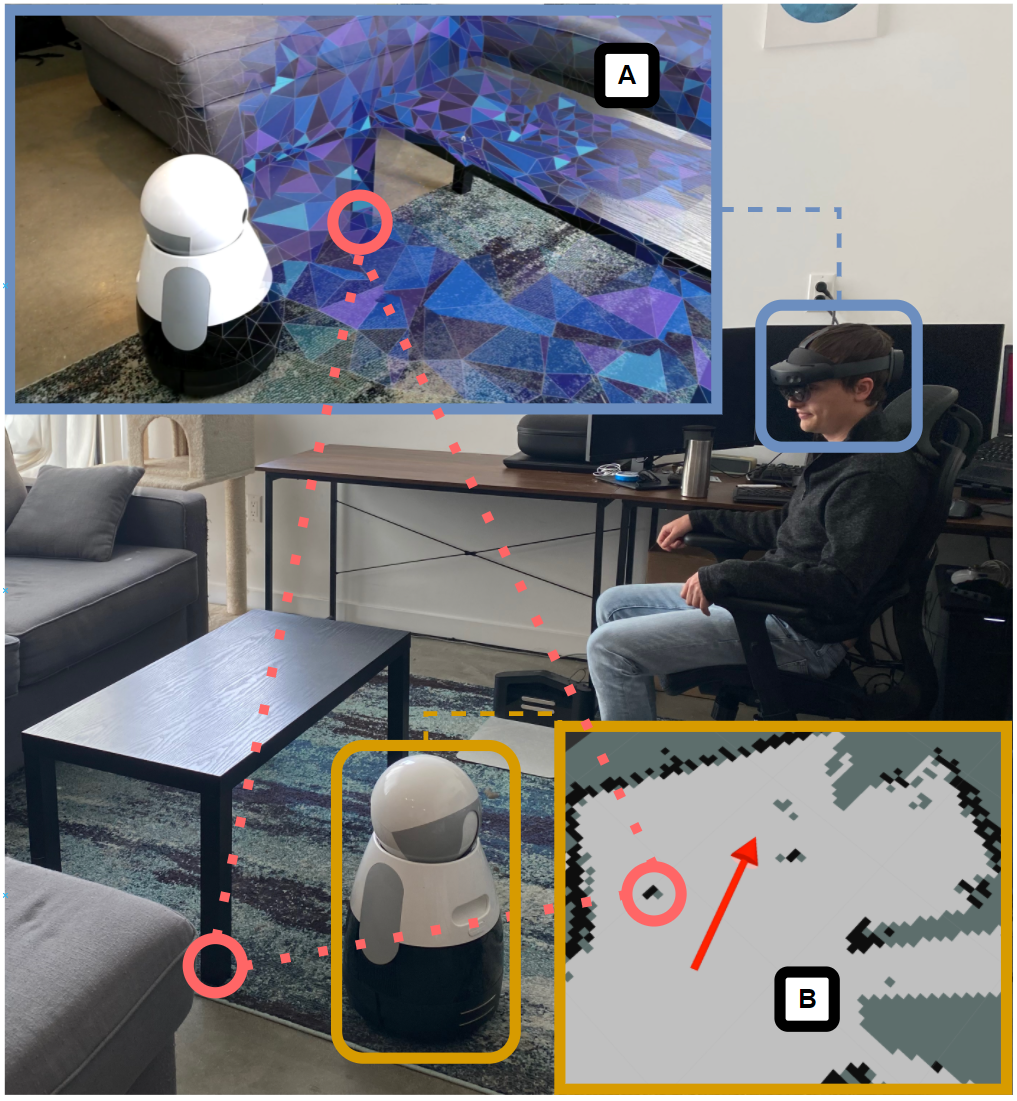}
\caption{Demonstrates a navigation situation where the robot 2D SLAM map (B) benefits from the 3D SLAM map from the ARHMD (A). The robot only maps the two front table legs (bottom left) as it is only equipped with a 2D lidar. The robot, however, is too tall to move past the table so it will collide if it does not use the 3D map from the ARHMD. A combined SLAM map would be created from feature matching such as the table legs (circles).}
\label{fig:vam-slam-model}
\end{figure}

\textit{Environment} - Data from the VAM technology further increases the robot's understanding of an environment. An example is provided in Fig. \ref{fig:vam-slam-model}. Given a mobile robot with 2D SLAM, a 3D map from an ARHMD's SLAM can be transformed into the robot's coordinate frame. The map can then be used for more accurate navigation. In another situation, a mobile phone camera can help with object recognition both in front and behind the robot.

\textit{User} - Data from VAM technology further increases the robot's understanding of the user. For example, a robot can better infer a user's intent to choose an object by using ARHMD eye-gaze \cite{rosenmixed}. Data gathered from motion sensors can be used \textcolor{black}{both for functional purposes (e.g., where is the human in relation to the robot) as well as} used to infer affective \textcolor{black}{human state} such as student curiosity \cite{groechel2021kinesthetic}.

\subsection{User Perceived Anchor Locations and Manipulability}
\label{sec:userperceivedanchor}
 \textcolor{black}{The mixed reality interaction design element (MRIDE) categorizations of user-anchored interface elements, environment-anchored interface elements, and virtual artifacts (described in {\ref{sec:mrides}}) are not mutually exclusive and lack necessary granularity.} For example, a virtual artifact can be user-anchored such as a movable user-anchored element or an environment anchored object that moves on its own. Granularity can also be added to benefit MRIDE classifications such as distinguishing between robot and environment anchored objects.

To this end, two important distinctions can be added to expand the current framework.  First, we apply two characteristics: \textbf{Anchor Location \{User, Robot, Environment\}} and \textbf{Perceived Manipulability \{User, Robot, None\}}.
 Second, we distinguish MRIDEs based on the intended user perception of the virtual object (i.e., where does the user perceive the anchor to be and who can/does move a virtual object). 
 
The first distinction allows for multiple labels within each characteristic, such as objects that are manipulable by both the robot and the user. Visuals for path planning (e.g., \cite{lemasurier2021semi}) further highlight the benefits of these granular distinctions. A planned robot pose visualized within the environment could be argued as both robot- and environment-anchored since the same trajectory can be defined within the robot's local frame of reference or within a global frame of reference.
 
The latter distinction is important \textcolor{black}{when characterizing Anchor Location} as any object can be translated into the environment's coordinate frame. \textcolor{black}{This translation may mathematically hold truth but the intended perception is important to the goals of studying a virtual object's effect on the user in the interaction. For example, the} granularity of Anchor Location combined with intended user perception allows for labeling virtual objects intended to be perceived as part of the robot such as adding virtual robot appendages \cite{tran2020exploring, groechel2019using}. These virtual \textcolor{black}{arms were} specifically designed to be perceived as part of the \textcolor{black}{robot to study their impact on the robot's functional and social expressivity, respectively. Therefore labeling the study of virtual arms as anchored to the environment or user does not help when grouping and looking for trends among different research projects.} 
 
\textcolor{black}{Further toward this idea,} a key property of virtual object manipulation is the user's action attribution of the manipulation (i.e., does the user perceive that they moved the object, the robot moved the object, or the object moved on its own). Perceived Manipulability is this action attribution, the perception the user has of the manipulation. For an object that the user manipulates (e.g., grabs), the Perceived Manipulability is the user. Virtual objects ``manipulated'' by the robotic system, however, are not necessarily directly manipulated by the robot nor perceived as so. In such a case, the virtual object may be scripted to move on its own to give the illusion of robot manipulation yet may fail in its illusion. When researching social robotics, this may have significant consequences on a user's perception of the robot (e.g., the robot's social presence). Therefore, to alleviate this complication and as stated above, TOKCS is applied from the intended user perception of the designed system (i.e., if the system attempts an illusion of robot manipulation of a virtual object, it is classified under \textit{Perceived Manipulability: Robot}).

Lastly, these \textcolor{black}{MRIDE} labels \textcolor{black}{are only applied to} virtual objects and are not tied to classifying VAM-HRI research under model, view, and control described in Sec.n \ref{sec:interactiondesignel} and \ref{sec::modelcomplexity}. VAM-HRI studies a variety of modalities provided by VAM technologies. HMD data used for improving a robot's SLAM, for example, still firmly sits under increasing the robot's internal complexity of model but is not applicable under Anchor Location nor Perceived Manipulability. \textcolor{black}{Thus these MRIDEs characteristics are designed for and only applied to virtual objects within VAM-HRI.}

\textcolor{black}{\subsection{Framework Limitations}
The TOKCS framework was designed to capture and classify the key characteristics of VAM-HRI systems at the time of writing. However, the framework may ultimately stand to be incomplete as advancements in both VAM-HRI research and VAM technology capabilities lead to currently nonexistent key characteristics differentiating VAM-HRI systems of the future. As field of VAM-HRI advances, the classification framework will likely need to grow as well.}

\section{Paper Classifications of the $4^{th}$ VAM-HRI Workshop}
\label{sec::paperclassifications}

TOKCS consists of characterizing VAM-HRI systems with the following: 

\begin{flushleft}

\textcolor{black}{\textbf{Anchor Location} \{User, Env, Robot\} -- where is the intended user perception of the virtual object's coordinate frame anchor (Sec {\ref{sec:userperceivedanchor}});}

\textcolor{black}{\textbf{Perceived Manipulability} \{User, Robot, None\} -- the intended user perception of ``who'' is able to or is currently manipulating the virtual object (Sec. {\ref{sec:userperceivedanchor}});}

\textcolor{black}{\textbf{Increases Expressivity of View (EV)} \{0,1\} -- VAM technology is used to more explicitly show a robot's internal model such as using virtual objects to visualize robot sensors (Sec. {\ref{sec:interactiondesignel}});}

\textcolor{black}{\textbf{Increases Flexibility of Controller (FC)} \{0,1\} -- using VAM technology to add control modality to a robot (Sec. {\ref{sec:interactiondesignel}});}

\textcolor{black}{\textbf{Increases Complexity of Model (CM)} \{0,1\} -- using VAM technology to help the robot's understanding of the environment and/or the interaction (Sec. {\ref{sec::modelcomplexity}}); }

\textcolor{black}{\textbf{Milgram Continuum \{AR, AV, VR\}} -- classification of which form of virtuality is being used (Sec. {\ref{sec:milgram}});}

\textcolor{black}{\textbf{Hardware Description} -- which VAM technology is used (Sec. {\ref{sec:hardware}})}

\textcolor{black}{\textbf{Software Description} --  which VAM software is used (Sec. {\ref{sec:software}})}

\end{flushleft}

\begin{table}[h!]
\caption{\label{tab:overall}Summary of TOKCS. Up arrow symbols ($\uparrow$) indicate that the work increases the functionality within this aspect of TOKCS. Blank entries indicate that the contributions of the paper for this aspect are on par with prior work.}
\tiny
 \begin{tabulary}{\linewidth}{LLLLLLLLL}
\toprule
Paper & Anchor Location & Perceived Manipulability &  Expressivity of View &  Flexibility of Controller  &  Complexity of Model & Milgram Continuum \citep{milgram1995augmented} & Software &                   Hardware \\
\midrule
\citet{Boateng2021-kd}  &                         Robot, Env &                                          &                          $\uparrow$ &                                &                          &                              AR &                               Unity &    Hololens video recordings via MTurk \\
\citet{ikeda2021ar}     &                                Env &                                         User &                          $\uparrow$ &                               $\uparrow$ &                         $\uparrow$ &                              AR &                               Unity &                               Hololens \\
\citet{lemasurier2021semi}         &                         Env, Robot &                                         User &                           &                               $\uparrow$ &                          &                              AV &                  Unity, ROSNET, ROS &                   HTC Vive \\
\citet{Puljiz2021-ue}             &                               &                                         &                           &                                &                         $\uparrow$ &                              AV &                               Unity &                               Hololens \\

Wadgao- nkar et. al [20] &                              Env, Robot &                                          &                          $\uparrow$ &                                &                          &                              AV &                               Unity &                               HTC VIVE \\
\citet{Barentine2021-lu}    &                                Env &                                         &                           &                               $\uparrow$ &                          &                              VR &                        Unity, TagUp &  Oculus Quest VR headset \& controllers \\
\citet{Higgins2021-ho}      &                               User &                                         User &                          $\uparrow$ &                               $\uparrow$ &                         $\uparrow$ &                              VR &            Unity, ROS$\#$, ROS, Gazebo &                       SteamVR headset \\

\citet{mara2021cobot}      &                                Env &                                  Robot, User &                           &                                &                          &                              VR &                               Unity &         HTC VIVE Pro Eye \& Leap Motion \\
\citet{Mimnaugh2021-ki}          &                                &                                         &                           &                                &                          &                              VR &                               Unity &                          Oculus Rift S \\
\citet{mott2021you}               &                               Env, User &                                          &                         $\uparrow$ &                                &                          &                              VR &                               Unity &                  MTurk Web Video of VR \\
\bottomrule
\end{tabulary}
\end{table}

We apply TOKCS to papers from the $4^{th}$ International Workshop on VAM-HRI to understand the ways in which researchers have been developing new techologies that leverage virtual, augmented, and mixed reality. The ten papers and their categorization within the TOKCS are summarized in Table \ref{tab:overall}. 

Within these ten papers, a variety of contributions were observed. In most cases, a given system focused its improvements on a specific dimension of the TOKCS; five of the ten papers developed improvements within a single dimension. The two that contributed expansions along all three axes leveraged AR/VR in a domain that had previously not utilized AR/VR. \citet{Higgins2021-ho} developed a method for training grounded-language models in VR, instead of with real world robots. \citet{ikeda2021ar} leverages AR-headsets for robotic debugging, where previous methods had used 2D screens. Four papers of the ten increased expressivity of view (EV), four increased the flexibility of the controller (FC), and three improved upon the robot internal complexity of model (CM). Of these papers, half can be described as virtual reality, three are augmented virtuality, and two are augmented reality. The majority of methods are anchored at the environment level. Two methods' anchor is located at the robot and two are located at the user. If a perceived manipulable is available, it is typically available at the user-level. 

We also observe a broad range of utilized hardware and software. Unity was overwhelmingly popular among papers as the 3D game engine of choice; nine of the ten papers explicitly mention Unity3D. The most popular HMD mentioned was the Hololens, which was used in three of the papers. Oculus Quest, HTC Vive, and MTurk are each used in two of the ten papers.

\subsection{Evaluations: Subjective and Objective Metrics}

In addtion to TOKCS, we further evaluated measures and metrics applied to VAM-HRI research. An important component of VAM-HRI research programs is to evaluate and benchmark new approaches by using both objective and subjective metrics. \textit{Objective} metrics are any metric that can be directly determined through sensors or measurements and do not involve a human's subjective experience. Examples of objective metrics include task completion time, the number of successful and failed trials, and accuracy and precision of visualization alignment.\textit{Subjective} metrics are any metric that depends on the perceived experience of the users involved. Examples of subjective metrics include mental workload, levels of immersiveness, and perceived system usability. Both subjective and objective metrics are important and complementary benchmarks for determining how effective new VAM-HRI contributions are compared to existing approaches. \textcolor{black}{A wide variety of metrics are available for these measurements,} and understanding which metrics VAM-HRI researchers are using helps highlight what aspects of interaction these technologies are improving on.

\begin{table}[h!]
\tiny
\caption{\label{tab:metrics}Description of objective and subjective metrics in 4th VAM-HRI Workshop papers. Blank spaces indicate a lack of metric of that type for that paper. Papers omitted from the table did not report metrics.}
\begin{tabulary}{\linewidth}{LLLL}
\toprule
Paper &                                        Objective Metrics & Subjective Metrics  \\ 
\midrule
\citet{Boateng2021-kd}           &           &            NASA TLX; Identification of robot position, orientation, and movement                                       \\
\citet{ikeda2021ar}              &     &                                System Usability Scale;
Think out loud process  \\
\citet{wadgaonkar2021exploring}          &                                       &  Post-experiment interviews; 
Custom survey questions      \\
\citet{Higgins2021-ho}             &          Task accuracy; Amount of training data          &                               
Custom survey questions \\
\citet{mara2021cobot}               &                                      Task completion time; Task completion rate &                                Custom survey questions  \\
\citet{Mimnaugh2021-ki}          &                     &                        Custom survey questions         \\
\citet{mott2021you}                 &                                     &                           Custom survey questions
\\

\bottomrule
\end{tabulary}
\end{table}

The most popular method of evaluating effectiveness of a given design was conducting surveys of study participants. Additional evaluation metrics focused on quantitative performance metrics on an evaluation task and subjective experience (see Table \ref{tab:metrics}). \textcolor{black}{Here we give general definitions for the categories of metrics used in the VAM-HRI contributions, and give examples from the contributions on how they implemented that metric for their application.}

\textcolor{black}{There were four objective metrics used in the VAM-HRI contributions: \textbf{Task accuracy}: the proportion of correct predictions to the total number of predictions (e.g: In \citet{Higgins2021-ho}, task accuracy is measured by the robot's ability to correctly classify the objects referred to by the human), \textbf{Amount of training data}: The amount of training data collected or is required for a machine learning application (e.g: In \citet{Higgins2021-ho}, the amount of training data refers to the amount necessary to close the sim2real gap versus learning-in-reality), \textbf{Task completion time}: the amount of time between tasks or events (e.g: In \cite{mara2021cobot}, the recorded time between robot signalling and human reaction),
\textbf{Task completion rate}:  The proportion of successful attempts at a task to the total number of attempts at the task (e.g: In \cite{mara2021cobot}, the number of successful completions of a minigame in a VR robot game environment).
}

\textcolor{black}{
There were 6 subjective metrics used in the VAM-HRI contributions: \textbf{NASA-task load index (NASA TLX) \cite{hart2006nasa}}: a multi-dimensional scale for measuring user workload during and after task execution (e.g: In \citet{Boateng2021-kd}, measuring user workload of situational awareness in proximal human-robot teaming with virtual shadows) , \textbf{Perceived robot identification}: user's perceived estimates about the robots in the environment (e.g:  In \citet{Boateng2021-kd}, users identified  what position, orientation and movement patterns of a out-of-sight robot member based on virtual shadows), \textbf{System Usability Scale (SUS) \cite{brooke1996sus}}: a questionnaire for measuring user's perceived usability of a system (fitness for purpose) on a seven-point Likert scale ranging from ``strongly disagree'' to ``strongly agree.'' (e.g: In \citet{ikeda2021ar}, SUS is used to assess the AR Robot debugging tool's usability), \textbf{Think out loud process}: participants actively voice their thoughts when using an application for researchers to receive real-time feedback (e.g:  In \citet{ikeda2021ar}, participants talk out-loud about their thought process when using the AR Robot debugging tool), \textbf{Interviews:} researchers ask participants to comment on specific features after using the VAM-HRI applications (e.g: In \citet{wadgaonkar2021exploring}, asking participants to comment on which robot features like color and texture impact  robot behavioral anthropomorphism in VR) \textbf{Custom survey questions}: similar to interviews, except users fill out specific custom survey questions that are application and task specific (e.g: In \citet{Higgins2021-ho}, users are asked about what they found frustrating for training ground language models in VR with simulated robots, or in \citet{Mimnaugh2021-ki} where users reported on VR sickness). 
}

\section{Current Trends \& the Future of VAM-HRI}
\label{sec::currenttrendsfuture}

In this paper, the $4^{th}$ VAM-HRI Workshop is used as a case study for MRIDE classification and categorization within the Reality Virtuality Interaction Cube; however, the papers submitted to this workshop can also be used to exemplify and project current and future trends in the field of VAM-HRI. This growing sub-field of HRI is showing promise in enhancing all areas of HRI from robot control (e.g., teleoperation and supervision interfaces) to collaborative robotics and improving teamwork with autonomous systems. The following will cover some of the key insights gathered from this year's workshop that show how VAM-HRI is evolving and improving the field of HRI as whole.

\begin{figure}[h]
\centering
\includegraphics[width=140mm]{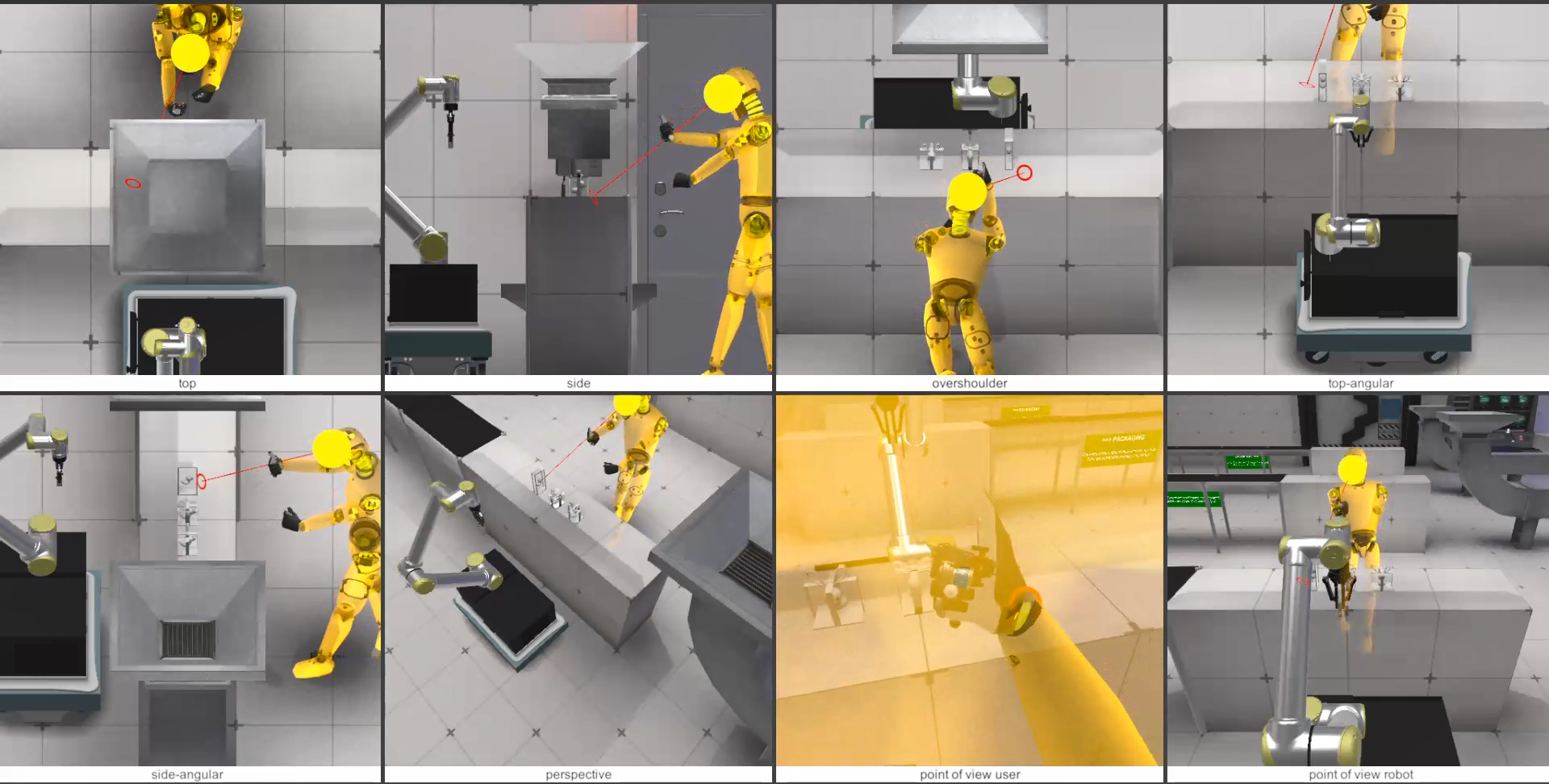}
\caption{\textcolor{black}{Advances in VAM-HRI research have enhanced the ability to precisely record, playback, and analyze human interactions with robots and other experimental stimuli in controlled user studies. This is exemplified in Mara et. al's \cite{mara2021cobot} CoBot Studio project where HRI user studies are conducted in a VR environment with numerous virtual cameras monitoring the experimental area from a multitude of angles. These cameras make use of the VR hardware to track body and head motion to record human postures and posture shifts, task-related human movements, gestures, and gaze behaviors, etc. Techniques such as this can benefit the field of HRI as a whole and allow for more complete and feature-rich data of human behavior that would otherwise be lost without VAM-HRI technology and recording techniques.}}
\label{fig:Cobot}
\end{figure}

\subsection{Experimental Evaluation of VAM-HRI Systems}

Research in HRI heavily features user studies in the evaluation of robotic systems and their interfaces. It has been an ongoing challenge to adequately record and playback human interactions with robot, to answer questions such as: `Where was the user looking at X time?,' `How close was the human positioned relative to the robot at Y moment?,' `What were the user's joint values when using a new interface and how are the physical ergonomics evaluated?' As a possible solution to many of these challenges, VAM-HRI allows for unprecedented recording, playback, and analysis of user interactions with virtual or real robots and objects in an experimental setting due to the inherent ability of HMDs (and other devices like a Leap Motion) to record body/hand/head position/orientation and gaze direction from a seemingly limitless number of virtual cameras recording from different angles \cite{williams2020using}. This is exemplified at a highly polished level in CoBot Studio \cite{mara2021cobot} (see Figure \ref{fig:Cobot}).

However, it is interesting to note that although precise objective measures can be relatively easily gathered from VAM-HRI experiments only 2 of the 10 submissions to the $4^{th}$ VAM-HRI Workshop gathered any objective data (see Table \ref{tab:metrics}). The lack of objective measures may be due to a handful of factors, such as the work being in a preliminary stage best suited for a workshop or the research questions being more focused on social responses and subjective opinions from users. Regardless of reason, we encourage authors of future VAM-HRI submissions to any venue to take full advantage of the objective measurements that VAM-HRI systems inherently provide, as objective observations are still useful for evaluating a multitude of social interactions (e.g., user pose for evaluating body language, user-robot proxemics, user gaze).

Although virtual reality interfaces have the aforementioned strengths for enhancing experimental evaluation, they have their own set of unique evaluation challenges as well, one of which being use of online studies with crowdworkers (e.g., on Amazon Mechanical Turk). HRI in general has made prolific use of online user studies (especially during the COVID-19 pandemic) that take advantage of cheap and readily available participants. However, VAM-HRI heavily draws upon 3D visualizations (as often seen in with HMD-based interfaces), which cannot be properly displayed to crowdworkers who lack HMDs and/or 3D monitors. Additionally, a strength of AR interfaces is that 3D data and visualizations can be rendered contextually in user's environments and are able to be observed from any angle desired by the user. \textcolor{black}{VAM-HRI studies that utilize crowdworkers to evaluate VAM interfaces, such as those performed by Mott et. al \cite{mott2021you}, are restricted to online images and videos viewed by Mechanical Turks on 2D monitors that restrict the user's viewpoint to that of pre-recorded videos which does not allow for a true VAM experience. It remains an open question if results from crowdsourced VAM-HRI studies provide comparable results to VAM-HRI studies run in person since 3D VAM technology is inherently experienced differently than the 2D experiences found on crowdsourcing platforms. Regardless, using crowdworkers still holds value in the early prototyping phases of VAM-HRI research where the initial formulation of object and interaction designs can be evaluated quickly and inexpensively.}

\subsection{VAM-HRI as an Interdisciplinary Study} 

HRI is well known to be an interdisciplinary field and VAM-HRI is showing to be no exception. The CoBot Studio project brings together roboticists, psychologists, AI experts, multi-modal communication researchers, VR developers, and professionals in interaction design and game design \cite{mara2021cobot}. As the VAM-HRI field grows, it will likely become increasingly common (and needed) to see teams with varied experiences and skill sets contributing to collaborative research.

\textcolor{black}{Research in multi-robot systems is an underexplored inspiration for VAM-HRI research in regard to enhancing the complexity of model (CM). VAM technology can be formulated as another robot within a system -- a robot with non-deterministic, non-directly controllable behavior but one with a data rich sensor suite.} The frameworks and techniques of the adjacent field may be able to be modified or even directly applied when treating the human user as an autonomous mobile sensor platform, akin to the human being treated as though they are another robot in the system. For example, spatial and semantic scene understanding are important perceptual capabilities for active robots (to navigate their environment) and passive VAM technologies (to localize the user's field of view).

Additionally, experimentation techniques seen in the field of general Virtual Reality may aid in the administering of questionnaires and gathering participant feedback. Typical questionnaires administered by VAM-HRI researchers can be quite jarring for participants who experience extreme context shifts between virtual worlds (where the study took place) and the real world (where the feedback is gathered). This poses as a potential confounding factor for participants who no longer visually reference what they are evaluating and may romanticize or incorrectly remember experimental stimuli they can no longer see. The field of Virtual Reality has similar challenges and some studies have started to provide in situ evaluations where questionnaires are posed to users within the virtual environments \cite{lin2019effect}. We are beginning to see this trend of in situ surveys in VAM-HRI as well. In the CoBot studio project, surveys are administered within the experiment's virtual setting, removing the confounding factors of: (1) reality-virtuality context shifts (having to leave the immersive virtual environment by taking off an HMD to take a mid-task survey); and (2) retrospective surveys provided well after exposure to experimental stimulus \cite{mara2021cobot}.

\textcolor{black}{The cross-disciplinary trends and ideas from the field of virtual reality are not unidirectional however;} VAM-HRI is currently posed to inform and improve the field of VR in return. Enhancing immersion has always been a primary goal of the field of VR since its inception many decades ago. With the rise of mass-produced consumer grade HMDs, visual immersion has reached new heights for users around the world. However, the challenge of providing \textit{physical} immersion through the use of haptics has largely remained an open question: how can a user reach out and touch a dynamic character in a virtual world? Research in VAM-HRI has proposed a potential solution for dynamic haptics, where robots mimic the pose and movements of virtual dynamic objects. Work by \citet{wadgaonkar2021exploring} exemplifies the notion of VAM-HRI supporting the field of VR with robots acting as dynamic haptic devices and allowing users to touch characters in virtual worlds and further enhance immersion in VR settings.

\subsection{Advancements in VAM-HRI}

A strength of VAM-HRI is the ability to alter a robot's morphology with virtual imagery. This technique can take the form of body extensions where virtual appendages are added to a real robot, such as limbs \cite{groechel2019using}, or form transformations where the robot's entire morphology is altered, such as transforming a drone into a floating eye \cite{walker2018communicating}. Recent VAM-HRI developments have further expanded upon this idea of changing a real robot's appearance through the aforementioned morphological alterations to include superficial alterations as well, where virtual imagery can be used to change a robot's cosmetic traits. Prior work has demonstrated that robot cosmetic alterations can communicate robot internal states (e.g., robotic system faults) \cite{de2018augmented}; however, to our knowledge, this is the first time such superficial alterations have been used to manipulate social interactions between human and robot \cite{wadgaonkar2021exploring}.

Although the interactions studied in HRI are typically focused on that of the end-user, a lesser studied category of interaction exists, which is that between robots and their developers and designers. Debugging robots often proves to be a challenging and tedious task with robot faults and unexpected behavior being hard to understand or explain without parsing through command lines and error logs. To address this issue, prior work in VAM-HRI has used AR interfaces to enhance debugging capabilities \cite{collett2010augmented, millard2018ardebug}. Work by \citet{ikeda2021ar} in VAM-HRI '21 has built upon these concepts by providing in situ AR visualizations of robot state and intentions, allowing users to better compare robots' plans with their actions when debugging autonomous robots. As AR hardware becomes increasingly intertwined with robotic systems, debugging tools such as these will likely become more commonplace to increase the efficiency and enjoyment of robot design.

Finally, VAM-HRI interfaces have been a popular topic of study within HRI for many years now, and many standard methods of interacting with robots through MR or VR have emerged (e.g., AR waypoints for navigation or AR lines for displaying robot trajectory \cite{walker2018communicating}). However, novel methods of interacting with robots are still being designed today, an example of which being persistent virtual shadows, aimed at tackling the issue of knowing a robot's location when out of the user's line-of-sight. Whereas prior solutions have tried using 2D top-down radars for showing robot locations \cite{walker2018communicating}, issues remain as interfaces such as these require repeated context shifts be performed by the user to look at the physical surroundings and then to the radar. Solutions such as persistent virtual shadows circumvent this limitation by embedding robot location data into the user's environment, providing a natural method of displaying a robot's location. This is a location cue that humans have learned to interpret almost subconsciously throughout the course of their lives. Creative advances such as these will continue to emerge in this relatively nascent sub-field of HRI, presenting an exciting new future for both VAM-HRI and the field of HRI as a whole.



\section*{Acknowledgments}
This work was supported by the National Science Foundation (NSF) under award IIS-1764092 and IIS-1925083. This work was also supported by the Draper Scholar Program. Any opinions, findings, and conclusions or recommendations expressed in this material are those of the authors and do not necessarily reflect the views of Draper.

\bibliographystyle{ACM-Reference-Format}
\bibliography{references}

\end{document}